\documentclass[sigconf]{acmart}
\AtBeginDocument{%
  }

\setcopyright{acmlicensed}
\copyrightyear{2026}
\acmYear{2026}
\acmDOI{XXXXXXX.XXXXXXX}
\acmConference[KDD '26]{32nd SIGKDD Conference on Knowledge Discovery and Data Mining}{Aug 09--13,
  2026}{Jeju, Korea}
\acmISBN{978-1-4503-XXXX-X/2018/06}

\usepackage{xcolor} 
\usepackage{soul}   
\usepackage{booktabs}
\usepackage{multirow}
\usepackage{graphicx} 
\usepackage{graphicx}       
\usepackage{subcaption}  
\usepackage[skip=0pt]{caption}

\begin{document}

\title{Decorrelating the Future: Joint Frequency Domain Learning for Spatio-temporal Forecasting}

\author{Zepu Wang}
\authornote{Both authors contributed equally to this research.}
\email{zepu@uw.edu}


\affiliation{%
  \institution{University of Washington}
  \city{Seattle}
  \state{WA}
  \country{USA}
}

\author{Bowen Liao}
\email{bliao6@asu.edu}

\authornotemark[1]

\affiliation{%
  \institution{Arizona State University}
  \city{Tempe}
  \state{AZ}
  \country{USA}
}

\author{Jeff (Xuegang) Ban}
\email{banx@uw.edu}
\affiliation{%
  \institution{University of Washington}
  \city{Seattle}
  \state{WA}
  \country{USA}
}

\renewcommand{\shortauthors}{Trovato et al.}

\begin{abstract}
Standard Direct Forecast models rely on point-wise objectives (e.g., Mean Squared Error) that overlook the complex spatio-temporal dependencies inherent in graph signals. While recent frequency-domain methods like FreDF mitigate temporal autocorrelation, they neglect spatial and cross-spatio-temporal structures. To address this, we propose FreST Loss (\textbf{Fre}quency-enhanced \textbf{S}patio-\textbf{T}emporal Loss), which extends supervision to the joint spatio-temporal spectrum. By leveraging the Joint Spatio-temporal Fourier Transform (\textit{JFT}), FreST Loss aligns predictions with ground truth in a unified spectral domain, effectively decorrelating complex dependencies. Theoretical analysis confirms that this approach reduces the estimation bias in time-domain training objectives. Extensive experiments on six real-world datasets demonstrate that FreST Loss is model-agnostic, consistently improving state-of-the-art baselines by capturing holistic spatio-temporal dynamics.
\end{abstract}




\maketitle

\section{Introduction}

Spatio-temporal forecasting is fundamental to numerous real-world applications, ranging from traffic flow prediction in Intelligent Transportation Systems~\citep{li2018diffusion, wang2023novel, wang2023st} to weather forecasting in meteorology~\citep{shi2015convolutional, lam2023learning}. In these critical domains, accurate and reliable forecasting is indispensable for optimizing resource allocation, enhancing public safety, and facilitating proactive decision-making.

Spatio-temporal forecasting utilizes historical observations to predict future spatio-temporal states, with methodologies generally categorized into Autoregressive and Direct Forecast (DF) paradigms. While autoregressive models generate predictions recursively and often suffer from error accumulation over long horizons \citep{li2018diffusion}, DF models predict the entire future sequence simultaneously in a single shot. This approach effectively circumvents recursive error propagation and enables efficient parallelization \citep{yu2018spatio}. In this work, we focus on the DF paradigm.

Within the DF paradigm, a substantial body of research has focused on designing sophisticated model architectures to effectively capture the intricate correlations embedded within historical inputs. To tackle this, diverse methodologies have been developed to model temporal autocorrelation, spatial correlations, and coupled cross-spatio-temporal dynamics. Early approaches predominantly utilized Recurrent Neural Networks (RNNs) and their variants (e.g., LSTM, GRU) to capture temporal dependencies by encoding historical sequences into evolving hidden states~\citep{hochreiter1997long, cho2014learning}. To explicitly account for spatial autocorrelation and topological constraints, Graph Neural Networks (GNNs) were subsequently introduced to propagate information among correlated nodes via message-passing mechanisms~\citep{kipf2016semi, wu2020connecting}. More recently, Transformer-based architectures have emerged as a powerful paradigm, leveraging self-attention mechanisms to capture global cross-spatio-temporal dependencies and long-range interactions simultaneously across both time and space~\citep{vaswani2017attention, xu2020spatial}.

However, beyond the complex dependencies in the historical input, the inherent correlations within the future spatio-temporal states are frequently overlooked. The vast majority of existing DF models employ point-wise objective functions, such as Mean Squared Error (MSE), to optimize model parameters~\citep{li2018diffusion, wu2020connecting}. From a probabilistic perspective, minimizing these objectives implicitly assumes that future observations are conditionally independent across distinct time steps and spatial nodes given the history. This independence assumption stands in stark contrast to the reality of spatio-temporal systems, where future states are highly correlated. For instance, traffic congestion propagates across adjacent nodes, and weather conditions persist over consecutive time steps. Consequently, point-wise optimization objectives fail to capture the joint conditional distribution of the future horizon, leading to suboptimal predictive performance.

Recently, FreDF~\citep{wang2025fredf} theoretically demonstrated that transforming the future states into the frequency domain effectively minimizes autocorrelation among time steps, as frequency components tend to be asymptotically independent. By optimizing in this orthogonal space, the model can substantially mitigate the bias inherent in standard time-domain objectives. However, this approach focuses exclusively on decoupling dependencies along the temporal dimension. It fundamentally overlooks spatial correlations among interconnected nodes and intricate cross-spatio-temporal dynamics, rendering it insufficient for modeling the joint distribution of complex graph signals.

To address this limitation, we introduce the \textbf{Frequency-enhanced spatio-temporal Loss (FreST Loss)}, a straightforward yet effective refinement for training DF models. The central idea is to align the forecasts and future states in the \textit{joint spatio-temporal frequency domain}, where the complex dependencies across both nodes and time steps are effectively diminished. This method resolves the theoretical discrepancy between the independence assumption of standard objectives and the entangled nature of graph signals, while retaining the advantages of the DF paradigm, such as sample efficiency and non-autoregressive simplicity.

Our main contributions are summarized as follows:
\begin{itemize}
    \item We identify spatio-temporal correlations (including spatial, temporal, and cross-correlations) as a critical yet underexplored challenge in spatio-temporal forecasting. We theoretically justify how minimizing standard time-domain errors introduces bias compared to the true negative log-likelihood (NLL).
    \item We propose FreST Loss, a novel objective function that transforms the optimization landscape into the joint frequency domain. By leveraging the asymptotic independence of spectral components, it effectively mitigates label autocorrelation and eliminates optimization bias. To our knowledge, this is the first effort to utilize joint frequency analysis to refine the learning objective in spatio-temporal forecasting.
    \item We validate the efficacy of FreST Loss through comprehensive experiments, demonstrating its universality and its ability to consistently enhance the performance of various state-of-the-art backbone models across diverse real-world datasets.
\end{itemize}

\section{Related Work}

\subsection{Spatio-temporal Forecasting}
Spatio-temporal data can be fundamentally formulated as a specialized instance of multivariate time series, where observations at different spatial locations correspond to distinct variable channels. Consequently, recent advances in general multivariate forecasting are often directly applicable to spatio-temporal tasks. State-of-the-art models have demonstrated remarkable efficacy on traffic and weather benchmarks by treating spatial points as multivariate channels. For instance, DLinear~\cite{zeng2023transformers} decomposes time series into trend and seasonal components, demonstrating that simple linear models can outperform complex Transformers when temporal dynamics are handled correctly. Similarly, iTransformer~\cite{liu2023itransformer} inverts the canonical Transformer architecture by applying attention and feed-forward networks across the variate dimension, effectively capturing global channel correlations. SparseTSF~\cite{lin2024sparsetsf} further pushes the efficiency boundary by utilizing sparse decoding techniques. These methods illustrate that capturing temporal dependencies and global channel interactions can yield competitive results even without explicit domain-specific spatial priors.

Distinct from general multivariate series, spatio-temporal data possesses inherent geometric relationships (e.g., road networks) that can be explicitly represented via graph structures. This topological information serves as strong prior knowledge for model design. Graph Neural Networks (GNNs) are widely adopted to leverage this structural bias. STGCN~\cite{yu2018spatio} integrates graph convolutions with temporal gated convolutions, using a pre-defined adjacency matrix to perform spatial message passing. Graph WaveNet~\cite{wu2019graph} extends this by proposing adaptive dependency matrices, allowing the model to capture latent spatial correlations beyond fixed graph structures. Beyond GNN-based approaches, recent works have explored efficient alternatives that bypass explicit graph propagation. STID~\cite{shao2022stid} employs simple Multi-Layer Perceptrons (MLPs) combined with spatial embeddings to capture local identities, while STAEformer~\cite{liu2023staeformer} utilizes specialized spatio-temporal attention mechanisms to model dynamic dependencies. These architectures focus heavily on designing effective encoders to extract coupled spatio-temporal representations.

Despite the architectural sophistication of the aforementioned methods—ranging from graph convolutions to attention mechanisms—the underlying training paradigm has remained largely stagnant. The vast majority of these models predominantly employ point-wise objective functions, such as Mean Squared Error (MSE). Theoretically, minimizing these point-wise losses implicitly assumes conditional independence among observations across both spatial and temporal dimensions (i.e., treating each space-time point as an isolated regression task). This assumption stands in direct contradiction to the inherent nature of spatio-temporal data, which is characterized by strong autocorrelation and complex spatial coupling. Consequently, while the encoders are designed to capture dependencies, the optimization objectives fail to account for the holistic correlation structure of future spatio-temporal states, potentially leading to suboptimal convergence and the oversmoothing of high-frequency details.



\subsection{Frequency Domain Methods in Forecasting}
Frequency domain analysis has emerged as a powerful paradigm in time series modeling. Pioneering works such as FEDformer~\cite{zhou2022fedformer} and Autoformer~\cite{wu2021autoformer} integrate Fourier decomposition within Transformer architectures to capture periodic patterns effectively. Building on this, FreTS~\cite{yi2023frets} demonstrates that frequency-domain MLPs act as superior learners by exploiting the energy compaction property of Fourier representations, while FourierGNN~\cite{yi2023fouriergnn} rethinks multivariate forecasting through a pure graph perspective in the spectral domain.

Recently, the application of frequency analysis has extended to the transformation of future states FreDF~\cite{wang2025fredf} demonstrates that transforming the future states into the temporal frequency domain effectively decorrelates time steps, thereby mitigating the bias inherent in DF objectives. By leveraging the asymptotic independence of frequency components for wide-sense stationary processes, FreDF aligns predictions with ground truth in the spectral domain, achieving significant improvements across diverse forecasting models. However, a critical limitation remains: FreDF focuses exclusively on temporal autocorrelation, neglecting the spatial dependencies that are fundamental to modeling complex spatio-temporal dynamics.




\section{Problem Formulation and Motivation}
In this section, we formally define the spatio-temporal forecasting problem and provide the theoretical motivation for our proposed approach.

\subsection{Spatio-Temporal Forecasting}
\subsubsection{Spatio-Temporal Signals.}
Let $\mathcal{G} = (\mathcal{V}, \mathcal{E})$ denote the underlying spatial structure, where $\mathcal{V} = \{1, \dots, N\}$ represents the set of $N$ nodes. We define the historical observations as a collection of scalars $\mathcal{X}_{hist} = \{ x_{t,n} \in \mathbb{R} \mid t \in \{1, \dots, T\}, n \in \{1, \dots, N\} \}$, where $x_{t,n}$ represents the feature value of node $n$ at historical time step $t$. Similarly, we define the ground-truth future sequence as $\mathcal{Y}_{true} = \{ y_{i,n} \in \mathbb{R} \mid i \in \{1, \dots, H\}, n \in \{1, \dots, N\} \}$, where $y_{i,n}$ denotes the value of node $n$ at the $i$-th future time step.

\subsubsection{Direct Forecasting Paradigm.}
In the \textit{Direct Forecast} (DF) paradigm, the model learns a mapping function $\mathcal{F}_\theta$ to project the historical context directly onto future spatio-temporal states:
\begin{equation}
    \hat{\mathcal{Y}} = \mathcal{F}_\theta(\mathcal{X}_{hist}),
\end{equation}
where $\hat{\mathcal{Y}} = \{ \hat{y}_{i,n} \}$ denotes the predicted spatio-temporal states.

\subsection{Limitation of DF and Motivation of FreST Loss}
\subsubsection{Optimization Objective and Limitations.}
Standard DF models are typically optimized by minimizing the MSE across all nodes and future time steps:
\begin{equation}
\label{eq:mse_loss}
    \mathcal{L}_{time} = \frac{1}{N \times H} \sum_{n=1}^{N} \sum_{i=1}^{H} (y_{i,n} - \hat{y}_{i,n})^2.
\end{equation}
While computationally efficient, this objective treats each prediction point $(i, n)$ in isolation. By doing so, it implicitly assumes conditional independence across both spatial and temporal dimensions. However, this assumption creates a significant theoretical gap in spatio-temporal tasks where data is inherently correlated. We rigorously analyze this discrepancy in the following theorem, extended from FreDF~\cite{wang2025fredf}.

\begin{theorem}[Bias of DF in the Spatio-Temporal Domain]
\label{thm:bias}
The learning objective of the standard DF paradigm (Eq. \ref{eq:mse_loss}) is biased with respect to the true negative log-likelihood (NLL) of the data distribution. By generalizing the analysis to both spatial and temporal dimensions, this bias can be explicitly decomposed as:

\begin{equation}
\label{eq:bias_st}
\begin{aligned}
\text{Bias} &= \underbrace{\sum_{n=1}^{N} \sum_{i=1}^{H} \frac{1}{2\sigma^2} (y_{i,n} - \hat{y}_{i,n})^2}_{\text{Standard MSE}} \\
&\quad - \underbrace{\sum_{n=1}^{N} \sum_{i=1}^{H} \frac{1}{2\sigma^2(1 - \mathcal{P}_{i,n}^2)} \left( y_{i,n} - \mu_{i,n|\text{pa}} \right)^2}_{\text{True spatio-temporal NLL}},
\end{aligned}
\end{equation}
where $\sigma$ denotes the standard deviation of the intrinsic noise (assumed to be constant). The term $\mu_{i,n|\text{pa}}$ represents the conditional mean given its spatio-temporal "parents" (past and neighboring dependencies). Here, $\rho_{(i,n),(j,u)}$ denotes the partial correlation between node $n$ at step $i$ and node $u$ at step $j$, and the cumulative correlation strength is defined as:
\begin{equation}
    \mathcal{P}_{i,n}^2 = \sum_{(j, u) \prec (i, n)} \rho_{(i,n), (j,u)}^2.
\end{equation}
\end{theorem}

The bias derived in Eq. \eqref{eq:bias_st} is strictly positive, stemming from the omitted correlation terms in the standard MSE. The term $\rho_{(i,n),(j,u)}$ effectively captures the complex joint distribution structure, which can be categorized into three distinct types:
\begin{enumerate}
    \item \textbf{Temporal Autocorrelation} ($n=u, i \neq j$): Captures the dependency of a node on its own historical residuals.
    \item \textbf{Spatial Correlation} ($n \neq u, i = j$): Captures the instantaneous dependency between different nodes (e.g., traffic conditions on adjacent roads).
    \item \textbf{Cross-Spatio-Temporal Correlation} ($n \neq u, i \neq j$): Captures wave-like propagation effects, where the past state of one node influences the future state of another.
\end{enumerate}

A critical challenge remains: \textit{How can we effectively incorporate the joint influence of these three types of correlations into the training process?} Addressing this limitation is the primary motivation for this work. To bridge the gap between the standard MSE objective and the spatio-temporal dependencies, we propose the FreST Loss.

\section{Discussion on Correlations in the Frequency Domain}

While FreDF~\citep{wang2025fredf} demonstrated that the Fast Fourier Transform (\textit{FFT}) can effectively mitigate autocorrelation bias in time series, spatio-temporal data presents a more complex challenge involving coupled dependencies. In this section, we systematically analyze these correlations in the frequency domain. We first review how \textit{FFT} handles temporal redundancies, then extend the paradigm to spatial structures via the Graph Fourier Transform (\textit{GFT}). Furthermore, we discuss the mitigation of cross-spatio-temporal correlations using the composition of \textit{FFT} and \textit{GFT}. This section concludes with empirical evidence demonstrating the superior performance of these frequency-domain techniques in handling spatio-temporal correlations.

\subsection{Decorrelating the Temporal Dimension}
We begin by addressing the temporal aspect. In time-series forecasting, historical values are often highly correlated with future values (autocorrelation), violating the independence assumption of standard regression losses. We utilize the Fast Fourier Transform (\textit{FFT}) to map these signals into an orthogonal basis.

\begin{definition}[Fast Fourier Transform]
\label{def:fft}
Let $\mathbf{y} \in \mathbb{R}^T$ be a univariate time series of length $T$. The Fast Fourier Transform (\textit{FFT}) projects $\mathbf{y}$ into the temporal frequency domain $\hat{\mathbf{y}} \in \mathbb{C}^T$ via the linear mapping:
\begin{equation}
    \hat{\mathbf{y}} = \mathit{FFT}(\mathbf{y}),
\end{equation}
where $\mathit{FFT}(\cdot)$ denotes the transformation operation based on the DFT matrix $\mathbf{F} \in \mathbb{C}^{T \times T}$ with entries $\mathbf{F}_{k,n} = e^{-j \frac{2\pi}{T} k n}$. 
\end{definition}

This transformation is not merely a change of representation; it fundamentally alters the statistical properties of the error surface, as described below.

\begin{theorem}[Temporal Decorrelation via \textit{FFT}]
\label{thm:temporal_decorrelation}
Let $\mathbf{y} \in \mathbb{R}^T$ denote a univariate time series. If $\mathbf{y}$ is a realization of a stationary process, its frequency components become asymptotically uncorrelated as the sequence length $T \to \infty$~\cite{wang2025fredf, brockwell1991time}. Specifically, the \textit{FFT} asymptotically diagonalizes the covariance matrix:
\begin{equation}
    \lim_{T \to \infty} \mathbb{E}[\hat{\mathbf{y}} \hat{\mathbf{y}}^\mathcal{H}] \approx \text{diag}(\mathbf{\Sigma}_{\text{time}}),
\end{equation}
where $\hat{\mathbf{y}} = \mathit{FFT}(\mathbf{y}) \in \mathbb{C}^T$ represents the signal in the frequency domain, $\mathcal{H}$ denotes the Hermitian transpose, and $\mathbf{\Sigma}_{\text{time}}$ represents the power spectral density (the eigenvalues of the temporal autocovariance).
\end{theorem}

By leveraging \textit{FFT}, temporal signals are projected into an orthogonal frequency basis where the autocovariance matrix is asymptotically diagonalized. This transformation effectively decorrelates temporal dependencies, transforming temporal autocorrelation into independent frequency components for more robust modeling.

\subsection{Decorrelating the Spatial Dimension}
Analogous to temporal autocorrelation, spatial graph signals exhibit strong inter-node dependencies dictated by the graph topology. Standard loss functions often ignore this geometric structure. To address this, we extend the decorrelation principle to the spatial domain using Graph Signal Processing (GSP) theory.

\begin{definition}[Graph Fourier Transform]
\label{def:gft}
Let $\mathcal{G} = (\mathcal{V}, \mathcal{E}, \mathbf{A})$ be a graph with $N$ nodes, and let $\mathbf{L}$ be its normalized or combinatorial Laplacian matrix. Let $\mathbf{L} = \mathbf{U} \mathbf{\Lambda} \mathbf{U}^\top$ be the spectral decomposition of the Laplacian, where $\mathbf{U} \in \mathbb{R}^{N \times N}$ is the matrix of orthonormal eigenvectors. The Graph Fourier Transform (\textit{GFT}) of a spatial signal $\mathbf{y}_s \in \mathbb{R}^N$ is defined as:
\begin{equation}
    \hat{\mathbf{y}}_s = \mathit{GFT}(\mathbf{y}_s) = \mathbf{U}^\top \mathbf{y}_s.
\end{equation}

\end{definition}

By projecting signals onto the eigenvectors of the Laplacian, we isolate spatial modes—ranging from global trends (low frequencies) to local variations (high frequencies)—which leads to the following decorrelation property.

\begin{theorem}[Spatial Decorrelation via \textit{GFT}]
\label{thm:spatial_decorrelation}
Let $\mathbf{y}_s \in \mathbb{R}^N$ denote a spatial signal observed on a graph $\mathcal{G}$. If $\mathbf{y}_s$ is a realization of a graph-stationary process~\cite{perraudin2017stationary, marques2017stationary}, its graph frequency components are uncorrelated. Specifically, the \textit{GFT} diagonalizes the spatial covariance matrix of the signal:
\begin{equation}
    \mathbb{E}[\hat{\mathbf{y}}_s \hat{\mathbf{y}}_s^\top] = \text{diag}(\mathbf{\Sigma}_{\text{space}}),
\end{equation}
where $\hat{\mathbf{y}}_s = \mathit{GFT}(\mathbf{y}_s) \in \mathbb{R}^N$ represents the signal in the graph spectral domain, and $\mathbf{\Sigma}_{\text{space}}$ represents the graph power spectral density.
\end{theorem}

By employing \textit{GFT}, spatial signals are projected onto the orthonormal eigenbasis of the graph Laplacian, effectively disentangling the geometric dependencies inherent in the graph topology. For graph-stationary processes, this transformation diagonalizes the spatial covariance matrix, thereby decorrelating the spatial dimensions and enabling the treatment of signals as independent spectral components.

\subsection{Decorrelating the Cross-Spatio-Temporal Dimension}
While \textit{FFT} and \textit{GFT} independently handle temporal and spatial redundancies, real-world spatio-temporal data involves complex \textit{cross-correlations}—such as "traffic waves," where congestion at a specific node propagates to its neighbors over subsequent time steps. Ideally, a robust forecasting model should decouple these intertwined dependencies. To achieve this, we propose utilizing the Joint Spatio-Temporal Fourier Transform (\textit{JFT}).

\begin{definition}[Joint Spatio-Temporal Fourier Transform]
\label{def:jft}
Let $\mathbf{Y} \in \mathbb{R}^{T \times N}$ represent a spatio-temporal signal matrix, where $T$ denotes the number of time steps and $N$ denotes the number of nodes in graph $\mathcal{G}$. The Joint Spatio-Temporal Fourier Transform (\textit{JFT}) projects $\mathbf{Y}$ into the joint spectral domain $\hat{\mathbf{Y}}_{joint} \in \mathbb{C}^{T \times N}$ via the composition of spectral transforms:
\begin{equation}
\label{eq:jft_commute}
    \hat{\mathbf{Y}}_{joint} = \mathit{JFT}(\mathbf{Y}) = \mathbf{F} \mathbf{Y} \mathbf{U},
\end{equation}
where $\mathbf{F} \in \mathbb{C}^{T \times T}$ is the temporal DFT matrix and $\mathbf{U} \in \mathbb{R}^{N \times N}$ is the spatial eigenvector matrix of the graph Laplacian. Due to the separability of these operators, the transform satisfies the commutativity property $\mathit{FFT}(\mathit{GFT}(\mathbf{Y})) = \mathit{GFT}(\mathit{FFT}(\mathbf{Y}))$, mapping the signal onto a unified basis formed by the Kronecker product of the temporal and spatial bases.
\end{definition}

The efficacy of this composite transform in removing cross-spatio-temporal correlations is rooted in the theory of \textit{Product Graphs} and \textit{Joint Stationarity}~\cite{loukas2016stationary, grassi2017time}. By modeling the spatio-temporal structure as a \textbf{Cartesian product graph} $\mathcal{G}_{st} = \mathcal{G}_{time} \square \mathcal{G}_{space}$, the signal can be viewed as a vectorized variable $\mathbf{y}_{vec} = \text{vec}(\mathbf{Y}) \in \mathbb{R}^{TN}$ with a joint covariance matrix $\mathbf{\Sigma}_{st} \in \mathbb{R}^{TN \times TN}$. Here, the operator $\square$ denotes the Cartesian product of graphs, which formally defines the joint topology where each node is connected to its temporal predecessors/successors and its spatial neighbors.

For a jointly stationary process~\cite{loukas2016stationary, grassi2017time}, the eigenvectors of the underlying product graph $\mathcal{G}_{st}$ are exactly the Kronecker product of the individual eigenvectors, $\mathbf{\Psi}_{st} = \mathbf{U} \otimes \mathbf{F}$. Consequently, the JFT defined in Eq.~\eqref{eq:jft_commute} is mathematically equivalent to projecting the signal onto this joint basis, which asymptotically diagonalizes the joint covariance matrix:
\begin{equation}
    \mathbb{E}[\text{vec}(\hat{\mathbf{Y}}_{joint})\text{vec}(\hat{\mathbf{Y}}_{joint})^\mathcal{H}] \approx \text{diag}(\mathbf{\Sigma}_{st}).
\end{equation}
This demonstrates that \textit{JFT} explicitly decouples the joint modes of variation, effectively neutralizing both temporal and spatial correlations simultaneously by exploiting the inherent Cartesian structure of the spatio-temporal domain.

\subsection{Empirical Analysis}
\label{sec:motivation}
To validate our theoretical analysis, we examine the correlation structure of real-world bike-sharing data (NYC-BIKE)~\cite{zhang2017deep} under different spectral transforms. The spatial graph is constructed via a thresholded Gaussian kernel over Euclidean distances between stations: $A_{ij} = \exp(-d_{ij}^2/\sigma^2)$ if $\exp(-d_{ij}^2/\sigma^2) \geq \epsilon$, and $0$ otherwise. Specifically, we measure the mean off-diagonal absolute correlation $\bar{\rho} = \frac{1}{n(n-1)}\sum_{i \neq j}|\rho_{ij}|$ across three perspectives: (1) \textbf{Temporal}: autocorrelation between distinct time steps; (2) \textbf{Spatial}: correlation between different sensor nodes; and (3) \textbf{Spatio-Temporal}: correlation within the flattened space-time representation. We compare four domain representations: the raw time domain, the \textit{FFT} domain, the \textit{GFT} domain, and the \textit{JFT} domain.

\begin{figure}[!t]
    \centering
    \includegraphics[width=0.8\linewidth]{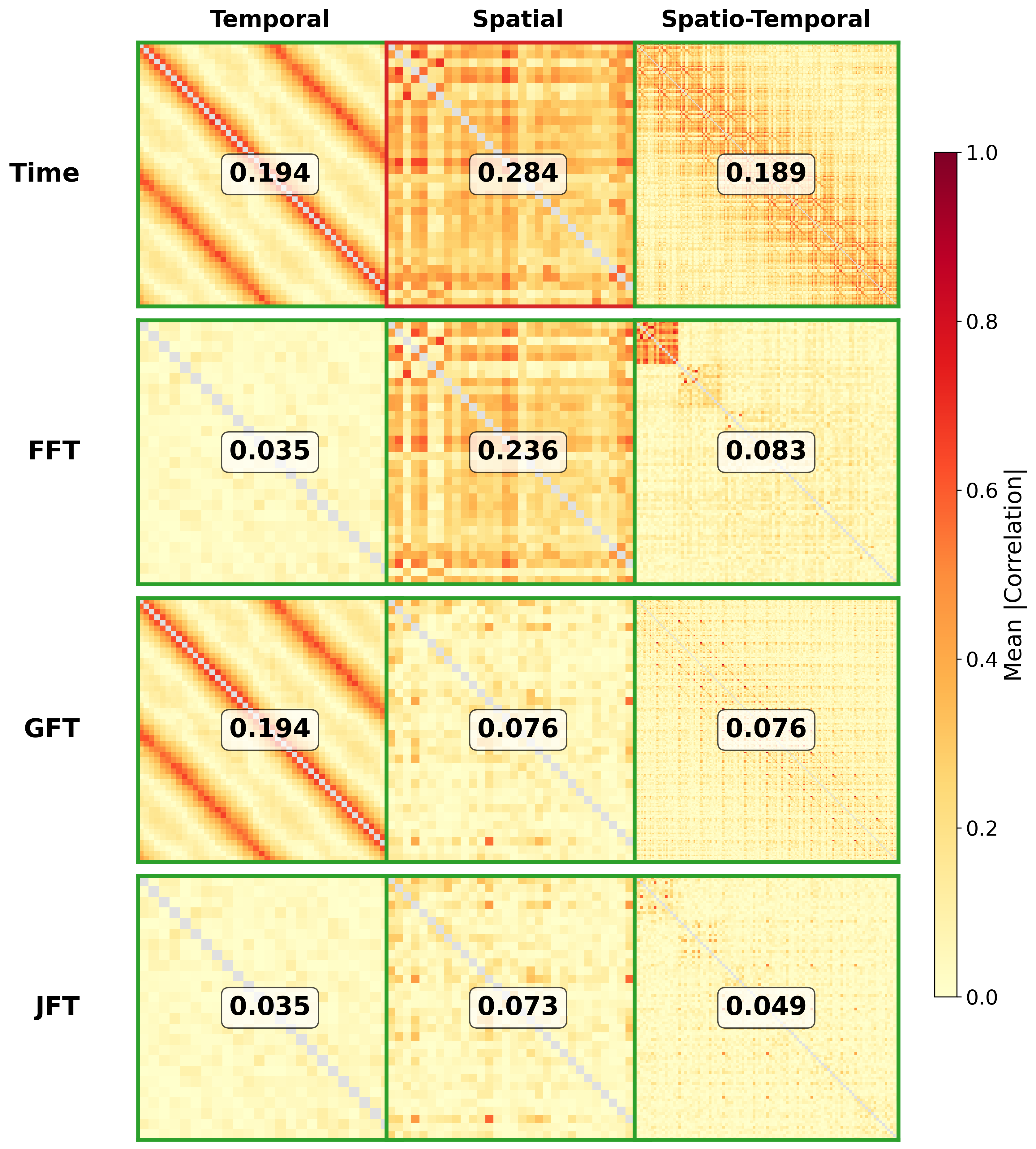}
    \caption{Correlation analysis of spatio-temporal data under different spectral transforms.}
    \label{fig:correlation}
\vspace{-5mm}
\end{figure}

As shown in Figure~\ref{fig:correlation}, the raw spatio-temporal signals exhibit substantial correlations across all three dimensions characterized by prominent structural patterns. The temporal correlation reaches $\bar{\rho}=0.194$, indicating strong autocorrelation between consecutive time steps. The spatial correlation is even more pronounced at $\bar{\rho}=0.284$, reflecting the inherent dependencies among neighboring sensor locations in the traffic network. The joint spatio-temporal correlation ($\bar{\rho}=0.189$) captures coupled dependencies where congestion propagates both across space and through time. These pervasive correlations violate the independence assumption underlying the standard MSE loss.

Figure~\ref{fig:correlation} reveals that neither \textit{FFT} nor \textit{GFT} alone can fully address this multi-dimensional correlation structure. \textit{FFT} effectively reduces temporal correlation ($0.194 \to 0.035$, an 82\% reduction) but leaves spatial correlation largely intact ($0.284 \to 0.236$, only a 17\% reduction). Conversely, \textit{GFT} reduces spatial correlation ($0.284 \to 0.076$, a 73\% reduction) but fails to address temporal dependencies ($0.194 \to 0.194$, a 0\% reduction). Notably, only \textit{JFT} achieves comprehensive decorrelation across all dimensions: temporal ($0.035$), spatial ($0.073$), and spatio-temporal ($0.049$)—the latter being significantly lower than either \textit{FFT} ($0.083$) or \textit{GFT} ($0.076$) alone. This empirical evidence confirms that \textit{JFT} is the only transform capable of simultaneously decoupling intertwined spatio-temporal dependencies, thereby providing a more orthogonal optimization landscape for forecasting models.

\begin{figure*}[t]
\caption{Overall Framework of FreST Loss. }
\begin{center}
\includegraphics[width=0.9\linewidth, height=6cm]{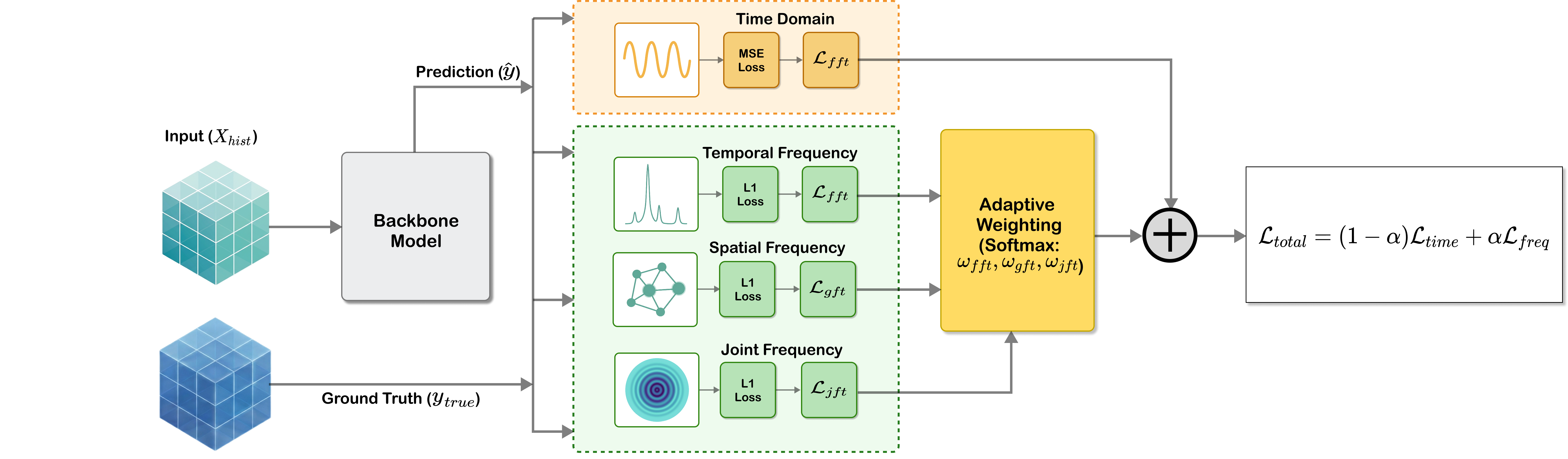}
\end{center}

\label{fig:model}
\end{figure*}

\section{Overall Framework}

To address the bias identified in Theorem \ref{thm:bias} and effectively decorrelate the predictive distributions, we propose the FreST Loss.
This proposed framework is \textbf{model-agnostic} and compatible with any DF architecture. Regardless of the backbone models, FreST Loss can be applied during the training phase to supervise the generated prediction $\hat{\mathcal{Y}}$ against the ground truth $\mathcal{Y}_{true}$.

The overall training objective is a weighted combination of time-domain fidelity and frequency-domain spectral consistency:
\begin{equation}
    \mathcal{L} = (1-\alpha)\mathcal{L}_{\mathrm{time}} + \alpha\mathcal{L}_{\mathrm{freq}},
\end{equation}
where $\alpha \in [0, 1]$ is a hyperparameter balancing point-wise accuracy and spectral structural alignment.

\subsection{Time Domain Loss}
The time domain component ensures the basic point-wise consistency between the prediction and the ground truth. We employ the standard Mean Squared Error (MSE) to minimize the error in the raw signal space:
\begin{equation}
    \mathcal{L}_{\mathrm{time}} = \frac{1}{N \times H} \|\mathcal{Y}_{true} - \hat{\mathcal{Y}}\|_F^2,
\end{equation}
where $\|\cdot\|_F$ denotes the Frobenius norm. 

\subsection{Frequency Domain Loss}
To explicitly capture and decorrelate the complex dependencies, we align the prediction $\hat{\mathcal{Y}}$ and ground truth $\mathcal{Y}_{true}$ in three complementary spectral domains.

We introduce three spectral discrepancy terms, each targeting a specific type of correlation:
\begin{itemize}
    \item \textbf{Temporal Consistency ($\mathcal{L}_{\mathrm{fft}}$):} Following [21], we apply \textit{FFT} along the time axis to capture temporal autocorrelation and periodic patterns.
    \item \textbf{Spatial Consistency ($\mathcal{L}_{\mathrm{gft}}$):} We utilize\textit{ GFT} based on the normalized Laplacian eigenvectors to capture spatial correlations and global topology.
    \item \textbf{Joint Spatio-temporal Consistency ($\mathcal{L}_{\mathrm{jft}}$):} To capture \textit{cross-spatio-temporal correlations} (e.g., propagation dynamics), we apply the composite transform \textit{JFT}.
\end{itemize}

It is worth noting that theoretically, the joint transform $\mathcal{JFT}$ encapsulates all spatio-temporal information, implying that $\mathcal{L}_{\mathrm{jft}}$ alone might suffice. However, in practice, optimizing solely on the joint spectrum presents significant engineering challenges, such as unstable convergence and the difficulty of disentangling coupled signals. Therefore, we explicitly retain $\mathcal{L}_{\mathrm{fft}}$ and $\mathcal{L}_{\mathrm{gft}}$ in our final objective. These terms provide direct supervision on the marginal distributions (pure temporal and pure spatial dependencies), serving as essential regularizers that stabilize the learning of the complex joint structure.

Specifically, different frequency components often exhibit vastly varying magnitudes, where lower frequencies typically possess significantly higher amplitudes compared to higher frequencies~\cite{wang2025fredf}. To ensure a more balanced and stable optimization process, the combined spectral loss components are defined using the $\ell_1$-norm, which also serves to encourage sparsity in the spectral representations:
\begin{equation}
\begin{aligned}
    \mathcal{L}_{\mathrm{fft}} &= \left\| \mathit{FFT}(\hat{\mathcal{Y}}) - \mathit{FFT}(\mathcal{Y}_{true}) \right\|_{1}, \\
    \mathcal{L}_{\mathrm{gft}} &= \left\| \mathit{GFT}(\hat{\mathcal{Y}}) - \mathit{GFT}(\mathcal{Y}_{true}) \right\|_{1}, \\
    \mathcal{L}_{\mathrm{jft}} &= \left\| \mathit{JFT}(\hat{\mathcal{Y}}) - \mathit{JFT}(\mathcal{Y}_{true}) \right\|_{1}.
\end{aligned}
\end{equation}

Since spectral energies in different domains often possess vastly different magnitudes, simple summation can lead to optimization dominance by a single term. To resolve this, we employ an adaptive mixing strategy. We first normalize each loss component by its detached magnitude ($\operatorname{stopgrad}$) to create scale-invariant gradients:
\begin{equation}
    \tilde{\mathcal{L}}_k = \frac{\mathcal{L}_k}{\operatorname{stopgrad}(\mathcal{L}_k) + \epsilon}, \quad \forall k \in \{\mathrm{fft}, \mathrm{gft}, \mathrm{jft}\}.
\end{equation}
We then combine these normalized losses using learnable weights $\mathbf{w}$ derived from a softmax-normalized parameter vector $\boldsymbol{\beta} \in \mathbb{R}^3$:
\begin{equation}
    \mathbf{w} = \operatorname{softmax}(\boldsymbol{\beta}), \quad \mathcal{L}_{\mathrm{freq}} = \sum_{k \in \{\mathrm{fft}, \mathrm{gft}, \mathrm{jft}\}} w_k \cdot \tilde{\mathcal{L}}_k.
\end{equation}
This mechanism allows the model to dynamically prioritize the most difficult spectral frequency components during the training process.

\section{Experiments}
\subsection{Setup}
\subsubsection{Datasets}

To comprehensively evaluate the generalization capability of our proposed framework across diverse spatio-temporal domains, we conduct experiments on six real-world datasets: NYC-Bike, AIR-BJ, AIR-GZ, METR-LA, PEMS-08, and SH-METRO. These datasets encompass three distinct categories of spatio-temporal tasks: 
(1) Traffic forecasting, including speed prediction (METR-LA~\cite{li2018diffusion}) and traffic flow prediction (PEMS-08~\cite{li2018diffusion}), which exhibit strong temporal periodicity; 
(2) Crowd flow/demand prediction (NYC-Bike~\cite{zhang2017deep} and SH-METRO~\citep{liu2020physical}), characterized by high variance; 
and (3) Environmental monitoring (AIR-BJ and AIR-GZ~\cite{yi2018deep}), reflecting the diffusion dynamics of PM$_{2.5}$ concentrations. 
Detailed statistics of these datasets are provided in Table~\ref{dataset}. 

For graph construction, adjacency matrices are built following the default settings from the original publications, tailored to the spatial characteristics of each domain. Specifically, traffic and bike-sharing datasets adopt a thresholded Gaussian kernel: $A_{ij} = \exp(-d_{ij}^2/\sigma^2)$ if $\exp(-d_{ij}^2/\sigma^2) \geq \epsilon$, and $0$ otherwise, where $d_{ij}$ denotes the pairwise distance, and $\sigma^2$, $\epsilon$ control the distribution and sparsity. The distance metric varies by dataset: road network distances for METR-LA, and Euclidean distances for PEMS-08 and NYC-Bike. Air quality datasets (AIR-BJ, AIR-GZ) employ the same Gaussian kernel with Haversine distances to account for geographical coordinates. For SH-METRO, a multi-graph structure is adopted, combining the physical metro topology with virtual graphs derived from DTW-based flow similarity and OD correlation patterns, among which we select the similarity graph for our implementation.

\begin{table}[t]
\caption{Statistics of datasets.}
\centering
\small
\setlength{\tabcolsep}{2pt}
\begin{tabular}{l c c c c l}
\toprule
Dataset & Domain & Nodes & Steps & Time Span & Graph Const. \\
\midrule
NYC-Bike & Bike & 250 & 1h & 04--09/2014 & Gaussian (Euc.) \\
AIR-BJ & Air & 35 & 1h & 01--12/2021 & Gaussian (Hav.)\\
AIR-GZ & Air & 41 & 1h & 01--12/2021 & Gaussian (Hav.)\\
METR-LA & Speed & 207 & 5min & 03--06/2012 & Gaussian (Road)\\
PEMS-08 & Flow & 170 & 5min & 07--08/2016 & Connectivity \\
SH-METRO & Metro & 288 & 15min & 07--09/2016 & Similarity \\
\bottomrule
\end{tabular}
\label{dataset}
\end{table}

\begin{table*}[!h]
\centering
\caption{Forecasting performance comparison. Improvements achieved by FreST loss relative to MSE Loss are marked in \textcolor{red}{red}, while performance degradation is marked in \textcolor{blue}{blue}. Note that DLinear, SparseTSF, and iTransformer are excluded from the PEMS-08 results as they do not yield competitive performance.}
\label{tab:deep_comparison}

\setlength{\tabcolsep}{1pt} 

\resizebox{\textwidth}{!}{
\begin{tabular}{l cccc cccc cccc cccc cccc cccc}
\toprule

\multirow{3}{*}{\textbf{Model}} 
 & \multicolumn{4}{c}{NYC-Bike} 
 & \multicolumn{4}{c}{AIR-BJ} 
 & \multicolumn{4}{c}{AIR-GZ} 
 & \multicolumn{4}{c}{METR-LA} 
 & \multicolumn{4}{c}{PEMS-08} 
 & \multicolumn{4}{c}{SH-METRO} \\

\cmidrule(lr){2-5} \cmidrule(lr){6-9} \cmidrule(lr){10-13} \cmidrule(lr){14-17} \cmidrule(lr){18-21} \cmidrule(lr){22-25}

 & \multicolumn{2}{c}{MSE Loss} & \multicolumn{2}{c}{FreST Loss} 
 & \multicolumn{2}{c}{MSE Loss} & \multicolumn{2}{c}{FreST Loss} 
 & \multicolumn{2}{c}{MSE Loss} & \multicolumn{2}{c}{FreST Loss} 
 & \multicolumn{2}{c}{MSE Loss} & \multicolumn{2}{c}{FreST Loss} 
 & \multicolumn{2}{c}{MSE Loss} & \multicolumn{2}{c}{FreST Loss} 
 & \multicolumn{2}{c}{MSE Loss} & \multicolumn{2}{c}{FreST Loss} \\

\cmidrule(lr){2-3} \cmidrule(lr){4-5} \cmidrule(lr){6-7} \cmidrule(lr){8-9} 
\cmidrule(lr){10-11} \cmidrule(lr){12-13} \cmidrule(lr){14-15} \cmidrule(lr){16-17} 
\cmidrule(lr){18-19} \cmidrule(lr){20-21} \cmidrule(lr){22-23} \cmidrule(lr){24-25}

& MAE & MSE & MAE & MSE 
& MAE & MSE & MAE & MSE 
& MAE & MSE & MAE & MSE 
& MAE & MSE & MAE & MSE 
& MAE & RMSE & MAE & RMSE 
& MAE & RMSE & MAE & RMSE \\
\midrule


STID & 
1.159 & 22.981 & \textcolor{red}{1.139} & \textcolor{red}{21.811} & 
0.265 & 0.127 & \textcolor{red}{0.255} & \textcolor{red}{0.124} & 
0.417 & 0.342 & \textcolor{red}{0.410} & \textcolor{red}{0.340} & 
0.468 & 0.486 & \textcolor{red}{0.455} & \textcolor{blue}{0.500} & 
42.573 & 62.418 & \textcolor{blue}{43.904} & \textcolor{red}{62.209} & 
91.636 & 230.135 & \textcolor{red}{81.525} & \textcolor{red}{187.214} \\ 

STGCN & 
1.302 & 30.092 & \textcolor{red}{1.288} & \textcolor{red}{30.123} & 
0.254 & 0.129 & \textcolor{red}{0.239} & \textcolor{red}{0.122} & 
0.398 & 0.334 & \textcolor{red}{0.379} & \textcolor{red}{0.307} & 
0.442 & 0.486 & \textcolor{red}{0.417} & \textcolor{red}{0.470} & 
34.219 & 54.567 & \textcolor{red}{32.072} & \textcolor{red}{49.615} & 
64.032 & 180.463 & \textcolor{red}{63.655} & \textcolor{blue}{181.125} \\ 

StemGNN & 
1.286 & 28.195 & \textcolor{red}{1.235} & \textcolor{red}{28.539} & 
0.414 & 0.245 & \textcolor{red}{0.381} & \textcolor{red}{0.218} & 
0.627 & 0.657 & \textcolor{red}{0.616} & \textcolor{blue}{0.658} & 
0.511 & 0.560 & \textcolor{red}{0.457} & \textcolor{red}{0.488} & 
40.097 & 78.742 & \textcolor{blue}{41.378} & \textcolor{red}{60.380} & 
87.506 & 187.705 & \textcolor{red}{71.887} & \textcolor{red}{154.875} \\ 

STDN & 
1.365 & 32.019 & \textcolor{red}{1.343} & \textcolor{red}{31.990} & 
0.233 & 0.130 & \textcolor{red}{0.207} & \textcolor{red}{0.105} & 
0.345 & 0.310 & \textcolor{red}{0.251} & \textcolor{red}{0.185} & 
0.438 & 0.555 & \textcolor{red}{0.408} & \textcolor{red}{0.527} & 
21.434 & 33.867 & \textcolor{red}{20.502} & \textcolor{red}{33.393} & 
83.849 & 217.015 & \textcolor{red}{78.689} & \textcolor{red}{206.341} \\ 

Staeformer & 
1.145 & 27.547 & \textcolor{blue}{1.183} & \textcolor{red}{27.543} & 
0.247 & 0.125 & \textcolor{red}{0.247} & \textcolor{red}{0.123} & 
0.416 & 0.343 & \textcolor{red}{0.414} & \textcolor{red}{0.340} & 
0.438 & 0.497 & \textcolor{red}{0.425} & \textcolor{red}{0.479} & 
33.373 & 47.504 & \textcolor{red}{29.551} & \textcolor{red}{41.810} & 
81.983 & 221.982 & \textcolor{red}{76.607} & \textcolor{red}{180.948} \\ 

DLinear & 
1.187 & 21.519 & \textcolor{red}{1.180} & \textcolor{red}{21.489} & 
0.271 & 0.133 & \textcolor{red}{0.256} & \textcolor{red}{0.127} & 
0.475 & 0.423 & \textcolor{red}{0.461} & \textcolor{red}{0.418} & 
0.504 & 0.529 & \textcolor{red}{0.491} & \textcolor{blue}{0.531} & 
/ & / & / & / & 
107.042 & 236.239 & \textcolor{red}{106.233} & \textcolor{blue}{236.295} \\ 

SparseTSF & 
1.258 & 21.638 & \textcolor{blue}{1.264} & \textcolor{red}{21.636} & 
0.275 & 0.137 & \textcolor{red}{0.269} & \textcolor{red}{0.136} & 
0.493 & 0.447 & \textcolor{red}{0.489} & \textcolor{red}{0.445} & 
0.561 & 0.722 & \textcolor{red}{0.559} & \textcolor{red}{0.719} & 
/ & / & / & / & 
146.734 & 308.735 & \textcolor{blue}{147.007} & \textcolor{red}{308.009} \\ 

iTransformer & 
1.120 & 20.292 & \textcolor{red}{1.111} & \textcolor{blue}{20.350} & 
0.262 & 0.134 & \textcolor{red}{0.243} & \textcolor{red}{0.125} & 
0.378 & 0.301 & \textcolor{blue}{0.379} & \textcolor{blue}{0.319} & 
0.516 & 0.649 & \textcolor{red}{0.500} & \textcolor{red}{0.637} & 
/ & / & / & / & 
71.802 & 176.744 & \textcolor{red}{66.596} & \textcolor{red}{160.290} \\ 

\bottomrule
\end{tabular}
}
\end{table*}
\subsection{Backbone Model Selections}
To comprehensively evaluate the universality of our proposed FreST Loss, we conduct experiments on eight state-of-the-art baseline models spanning three dominant paradigms in spatio-temporal forecasting: 
(1) \textbf{Linear/MLP-based models}, including STID~\cite{shao2022stid} specialized for spatio-temporal tasks, alongside DLinear~\cite{zeng2023transformers} and SparseTSF~\cite{lin2024sparsetsf} for general multivariate time series (MTS) forecasting; 
(2) \textbf{Transformer-based architectures}, represented by the spatio-temporal model STAEformer~\cite{liu2023staeformer} and the MTS-oriented iTransformer~\cite{liu2023itransformer}; 
and (3) \textbf{Spatio-Temporal Graph Neural Networks (ST-GNNs)}, specifically STGCN~\cite{yu2018spatio} and STDN~\cite{cao2025spatiotemporal} for explicit spatio-temporal modeling, and StemGNN~\cite{cao2020stemgnn} for graph-based MTS forecasting.


\subsubsection{Implementation Details}
We implement all baseline models and our proposed framework based on BasicTS~\citep{shao2024exploring}. The historical input length $T$ and the prediction horizon $H$ are set to 48 and 96, respectively. By leveraging the unified data pipelines and model configurations provided by BasicTS, we ensure consistent experimental settings across diverse architectures to minimize implementation bias. To strictly evaluate the effectiveness of our proposed loss, we conduct controlled experiments: for each backbone model, we maintain identical hyperparameters (e.g., learning rate, batch size, random seed) when training with the standard MSE loss versus our proposed FreST Loss. All experiments are conducted on a server equipped with a single NVIDIA GeForce RTX 4090 GPU. The balancing hyperparameter $\alpha$, which controls the weight of the frequency-domain loss, is set to $0.5$ for all experiments except for SH-METRO, where $\alpha = 0.9$; additionally, we provide sensitivity analysis of $\alpha$ in Section~\ref{sec:alpha_ablation}.

\subsection{Experiment Analysis}
We first evaluate the consistent performance gains provided by FreST Loss across various backbone models. Subsequently, we provide a multifaceted analysis of FreST's performance from several perspectives.

\subsubsection{Overall Result}
Table~\ref{tab:deep_comparison} presents a comprehensive performance comparison across six diverse datasets. The results consistently demonstrate the superiority of the proposed FreST Loss (Ours) over the standard MSE Loss. Across the 44 total metrics reported (excluding null entries), our method achieves performance improvements in 88.6\% of the cases (39 out of 44), confirming its stability and generalizability across different neural architectures and spatio-temporal patterns.

The magnitude of improvement is particularly significant in complex urban datasets. For instance, in the SH-METRO dataset, the MAE for StemGNN is reduced from 87.506 to 71.887, representing a substantial 17.8\% improvement. Similarly, in AIR-GZ, the STDN model sees its MAE drop from 0.345 to 0.251, a remarkable 27.2\% reduction. These significant gains highlight that integrating joint spatio-temporal frequency domain supervision allows models to better capture intrinsic periodicities and dependencies that traditional time-domain losses often overlook. The occasional marginal degradations merit further investigation into the interplay between graph structure and spectral supervision.

\subsubsection{Ablation Study of JFT Components}

In this study, we evaluate the performance of various frequency-based loss components, with the results summarized in Table~\ref{tab:jft_comparison}. It is observed that all frequency-based losses consistently achieve better performance than the standard MSE baseline. This improvement underscores the importance of spatio-temporal decorrelation during neural network training. Notably, our proposed FreST Loss demonstrates superior robustness and precision, ranking among the top two across 87.5\% of the evaluated metrics. This exceptional performance confirms the effectiveness of FreST Loss in capturing complex dependencies for real-world applications.
\begin{table}[t]
\centering
\caption{Comparison of Ablation Results for STGCN and DLinear on AIR-GZ and METR-LA Datasets. Best results are in \textcolor{red}{red}, and second best are in \textcolor{orange}{orange}.}
\label{tab:jft_comparison}
\renewcommand{\arraystretch}{1} 
\setlength{\tabcolsep}{4pt}    
\small
\begin{tabular}{l cccc cccc}
\toprule
\multirow{2}{*}{Method} & \multicolumn{4}{c}{STGCN} & \multicolumn{4}{c}{DLinear} \\
\cmidrule(lr){2-5} \cmidrule(lr){6-9}
& \multicolumn{2}{c}{AIR-GZ} & \multicolumn{2}{c}{METR-LA} & \multicolumn{2}{c}{AIR-GZ} & \multicolumn{2}{c}{METR-LA} \\
\cmidrule(lr){2-3} \cmidrule(lr){4-5} \cmidrule(lr){6-7} \cmidrule(lr){8-9}
& MAE & MSE & MAE & MSE & MAE & MSE & MAE & MSE \\
\midrule
MSE & 0.398 & 0.334 & 0.442 & 0.486 & 0.475 & 0.423 & 0.504 & 0.529 \\
\textit{FFT} & 0.398 & 0.335 & 0.434 & 0.489 & 0.462 & \textcolor{orange}{0.418} & 0.499 & \textcolor{orange}{0.523} \\
\textit{GFT} & 0.385 & 0.321 & 0.440 & 0.484 & 0.458 & 0.420 & 0.507 & \textcolor{red}{0.522} \\
\textit{JFT} & 0.383 & \textcolor{orange}{0.318} & 0.419 & \textcolor{orange}{0.471} & 0.471 & \textcolor{orange}{0.418} & 0.504 & \textcolor{red}{0.522} \\
\textit{FFT}+\textit{GFT} & \textcolor{red}{0.375} & 0.319 & \textcolor{orange}{0.420} & 0.477 & \textcolor{red}{0.453} & 0.420 & \textcolor{orange}{0.493} & 0.526 \\
FreST Loss & \textcolor{orange}{0.379} & \textcolor{red}{0.307} & \textcolor{red}{0.417} & \textcolor{red}{0.470} & \textcolor{orange}{0.461} & \textcolor{red}{0.418} & \textcolor{red}{0.491} & 0.531 \\
\bottomrule
\end{tabular}
\end{table}
\subsubsection{Improvement versus various future sequence length}
\begin{figure}[htbp]
 \caption{Performance comparison of STGCN in AIR-BJ and iTransformer in Air-GZ with various future sequence lengths}
    \centering
    \begin{subfigure}[b]{0.22\textwidth}
    \caption{MAE (STGCN) in AIR-BJ}
        \centering
        \includegraphics[width=\textwidth]{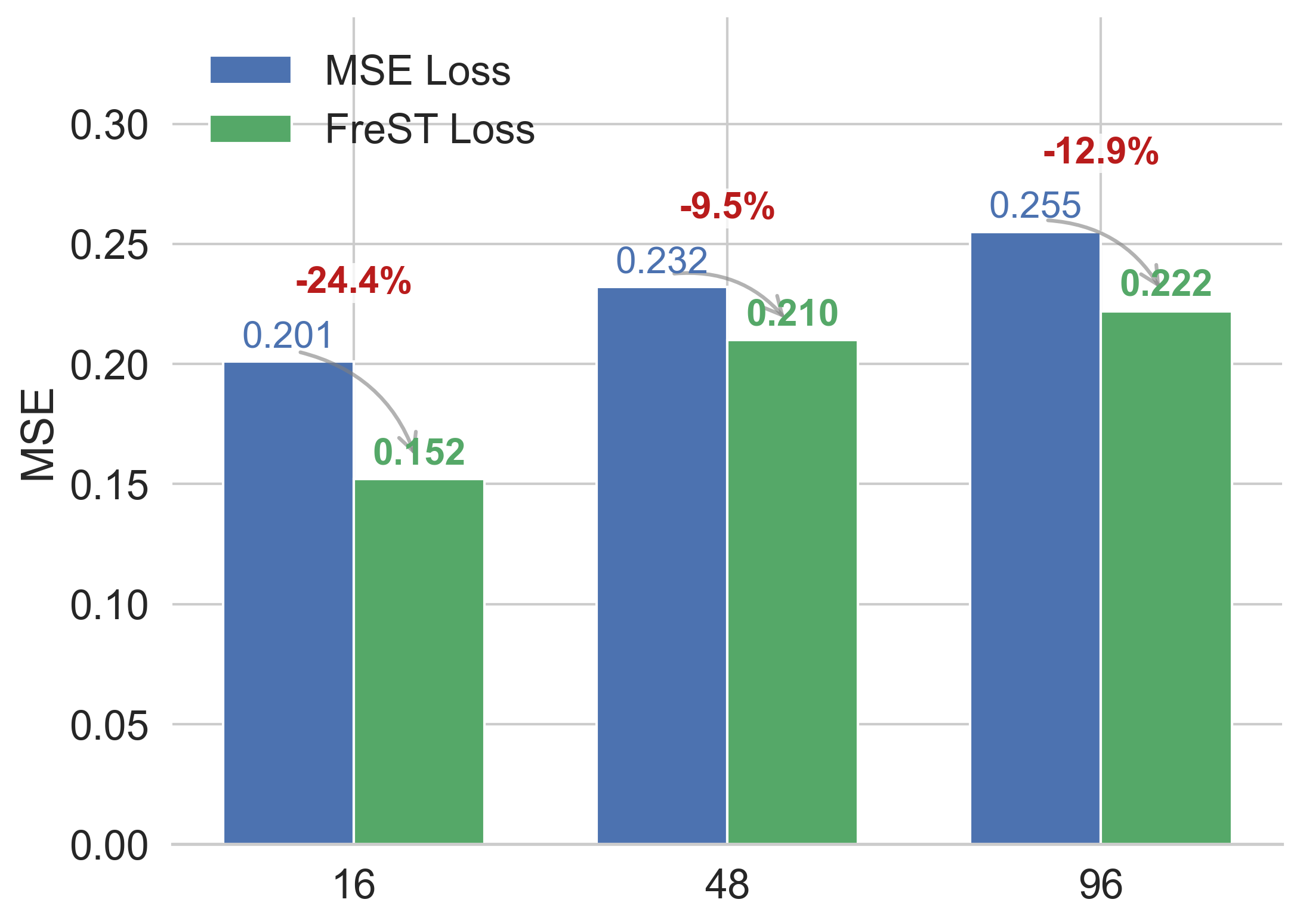}
        
        \label{fig:mae}
    \end{subfigure}
    \hfill 
    \begin{subfigure}[b]{0.22\textwidth}
    \caption{MSE (STGCN) in AIR-BJ}
        \centering
        \includegraphics[width=\textwidth]{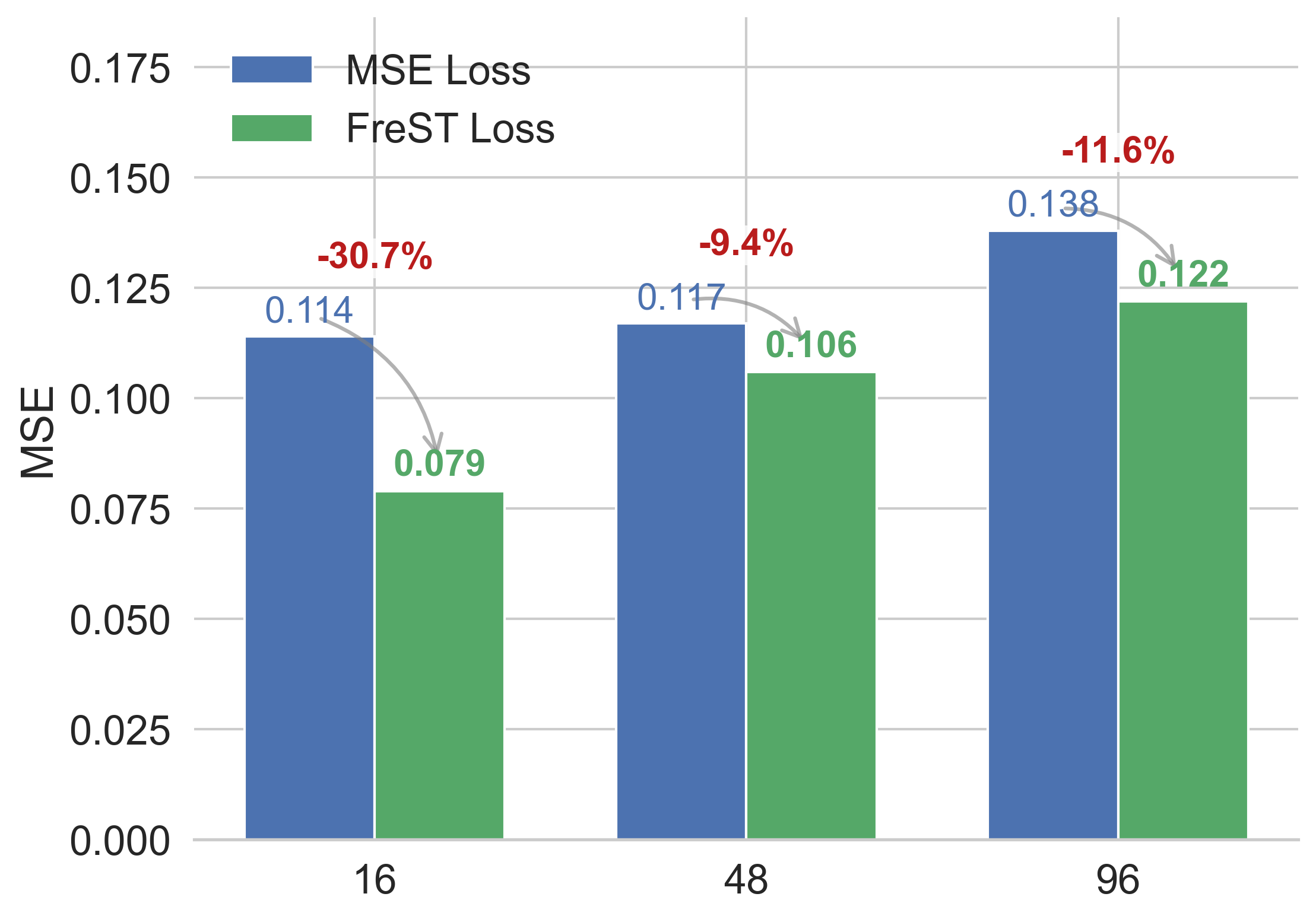}
        
        \label{fig:mse}
    \end{subfigure}
     \begin{subfigure}[b]{0.22\textwidth}
     \caption{MAE (iTransformer) in AIR-GZ}
        \centering
        \includegraphics[width=\textwidth]{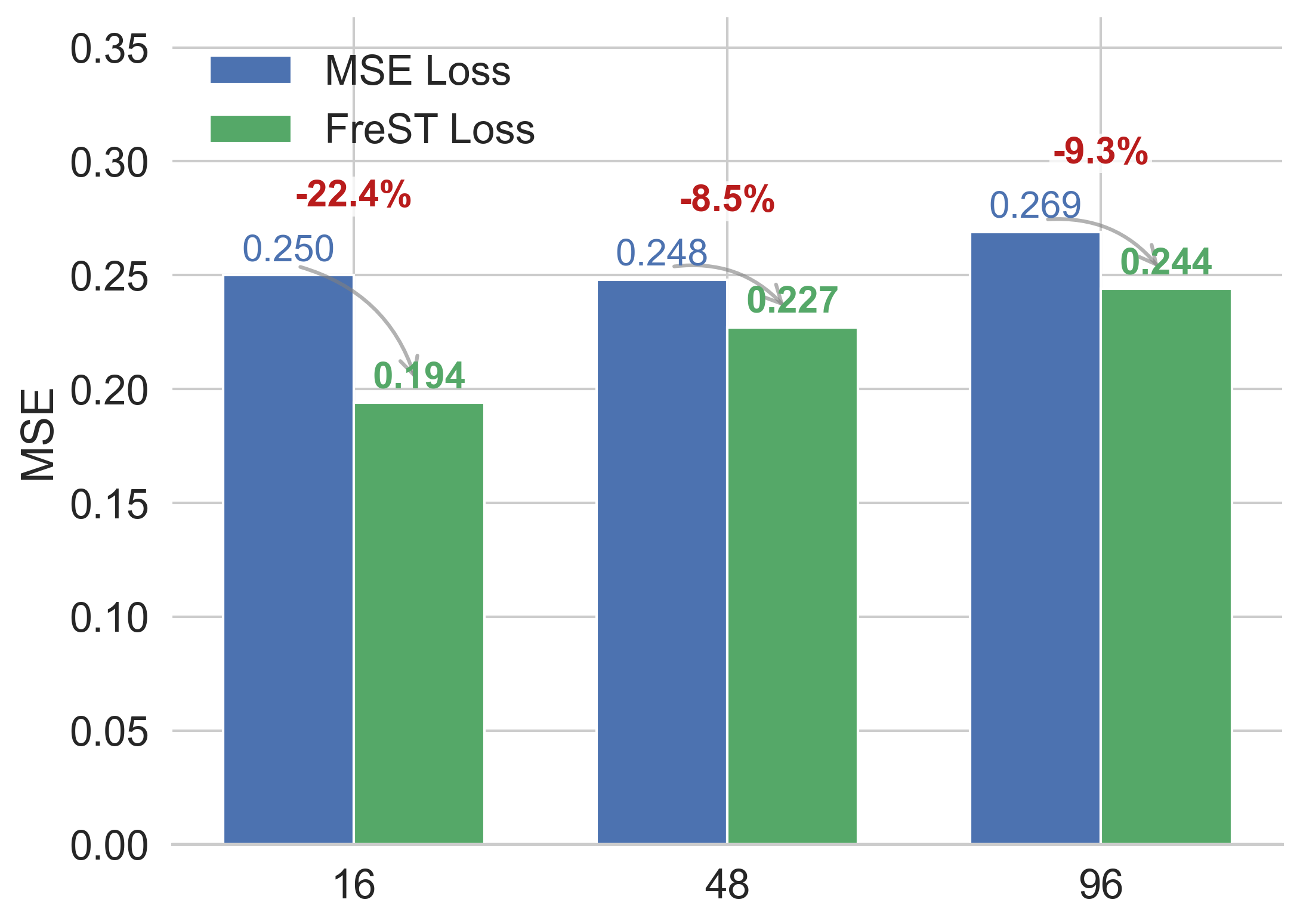}
        
        \label{fig:mse}
    \end{subfigure}
    \hfill 
     \begin{subfigure}[b]{0.22\textwidth}
     \caption{MSE (iTransformer) in AIR-GZ}
        \centering
        \includegraphics[width=\textwidth]{figures/iTransformer_mse_GZ.png}
        
        \label{fig:mse}
    \end{subfigure}

   
    \label{fig:comparison}
\end{figure}

To validate the effectiveness of FreST Loss across varying prediction horizons, we evaluate three different future sequence lengths, as shown in Figure \ref{fig:comparison}. FreST Loss consistently enhances performance across all models, with particularly significant gains (exceeding 20\%) observed in shorter sequences. This is because temporal autocorrelation is significantly stronger in the short term, which typically introduces a heavy optimization bias in traditional losses. By theoretically decoupling these dependencies in the joint spectral domain, FreST effectively removes this autocorrelation bias.

\subsubsection{Generalization Capacity of FreST}
\begin{figure}[htbp]
 \caption{Training and Validation RMSE Curve for Staeformer in SH-METRO}
    \centering
    \begin{subfigure}[b]{0.22\textwidth}
     \caption{Training Curve}
        \centering
        \includegraphics[width=\textwidth]{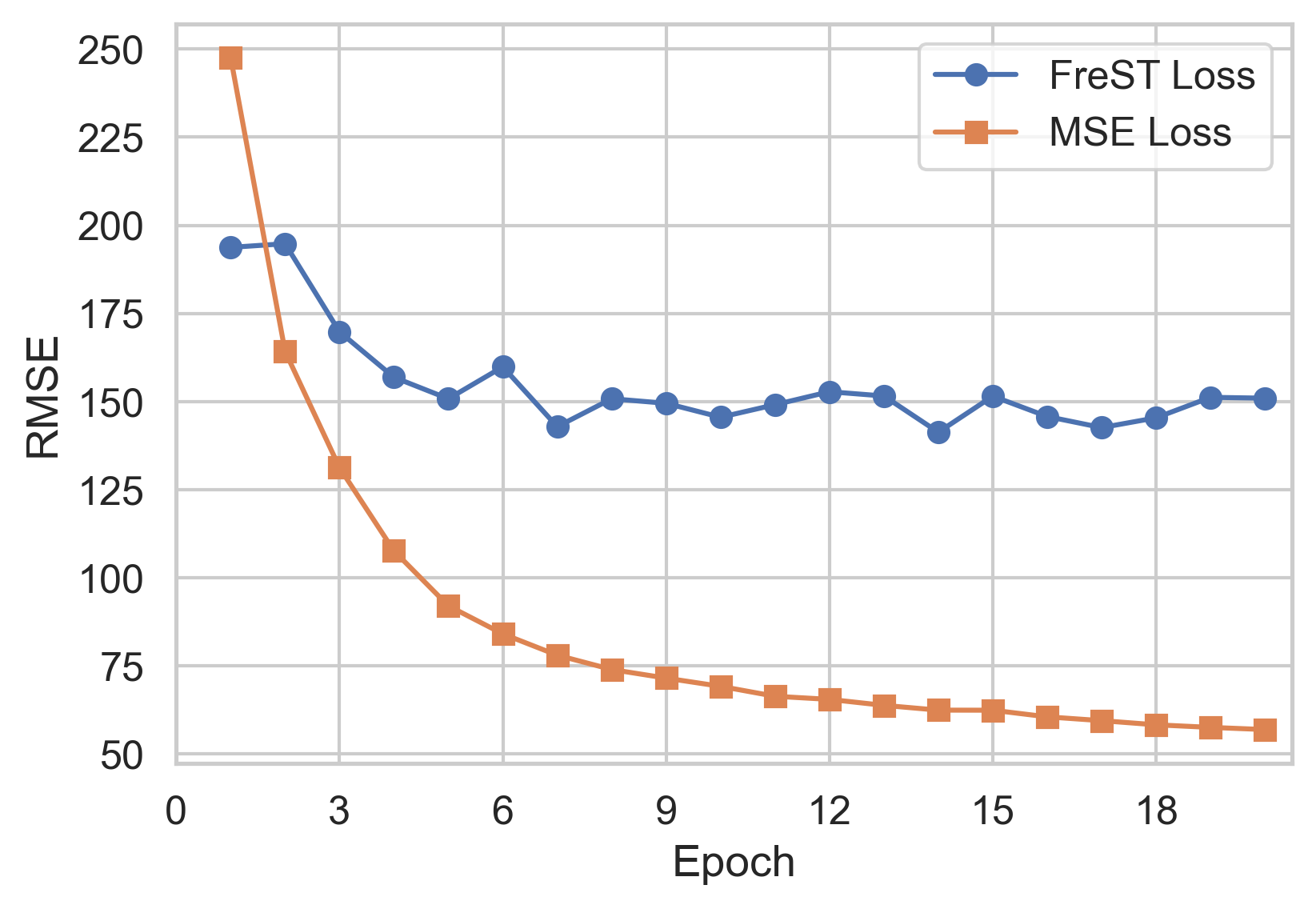}
       
        \label{curve_rmse_train}
    \end{subfigure}
    \hfill 
    \begin{subfigure}[b]{0.22\textwidth}
    \caption{Validation Curve}
        \centering
        \includegraphics[width=\textwidth]{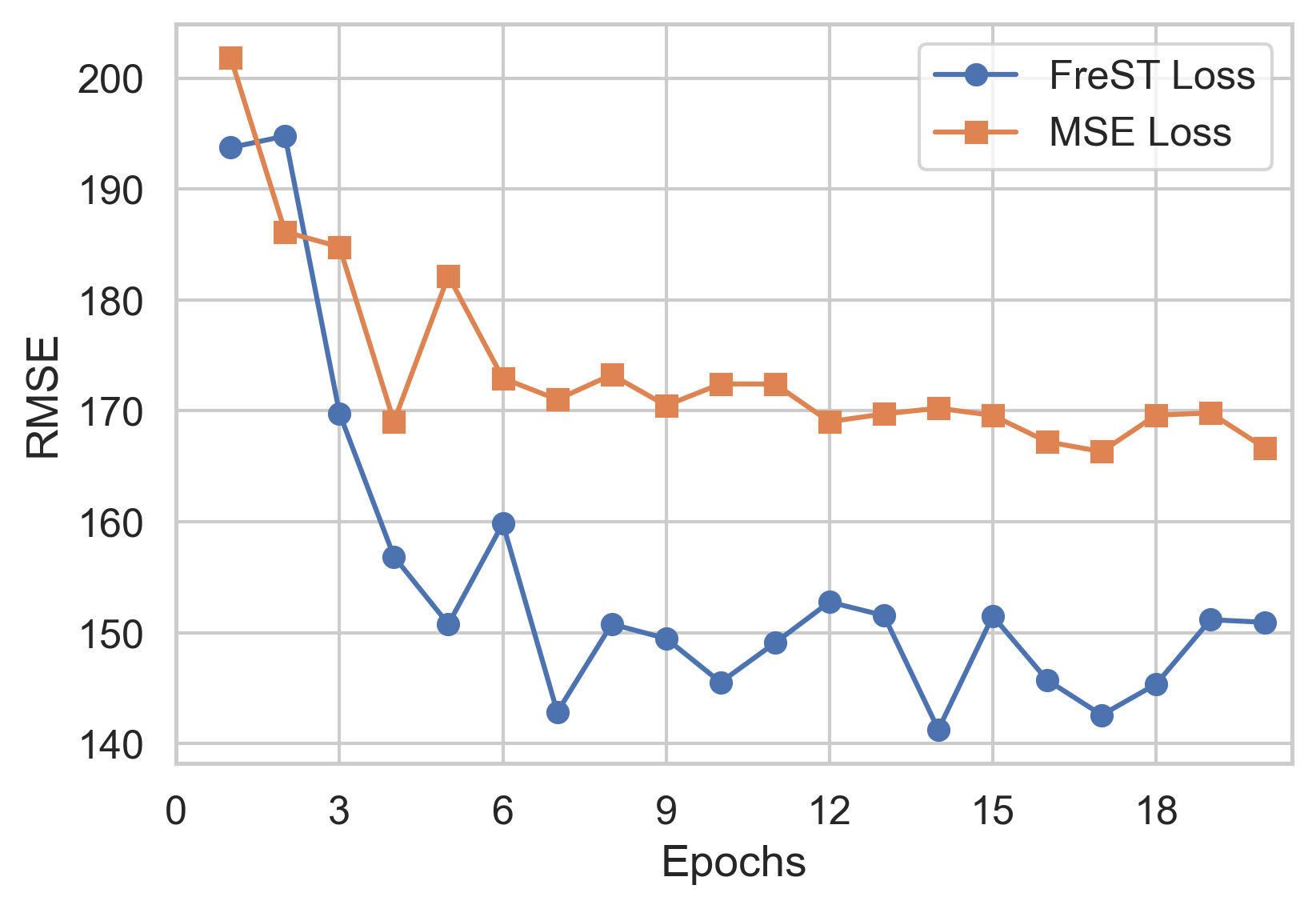}
        
        \label{curve_rmse_val}
    \end{subfigure}

   
    \label{training curve}
\end{figure}

The learning curves presented in Figure~\ref{training curve} highlight a fundamental difference in convergence behavior between the two objectives. While standard MSE optimization achieves minimal residuals on the training set, it suffers from a substantial generalization gap, evidenced by elevated and erratic errors on the testing set. This discrepancy suggests that MSE drives the model to memorize transient noise and specific data artifacts within the training distribution rather than learning transferable laws. Conversely, FreST Loss exhibits a much tighter alignment between training and testing performance. By constraining the optimization to the joint spatio-temporal frequency domain, FreST acts as an implicit regularizer: it effectively filters out non-generalizable high-frequency noise and forces the model to focus on invariant structural patterns.

\subsubsection{Importance of $\alpha$}
\label{sec:alpha_ablation}

\begin{figure}[htbp]
\caption{Impact of $\alpha$ on Model Performance in AIR-BJ}
    \centering
    \begin{subfigure}[b]{0.23\textwidth}
    \caption{STGCN for AIR-BJ}
        \centering
        \includegraphics[width=\textwidth]{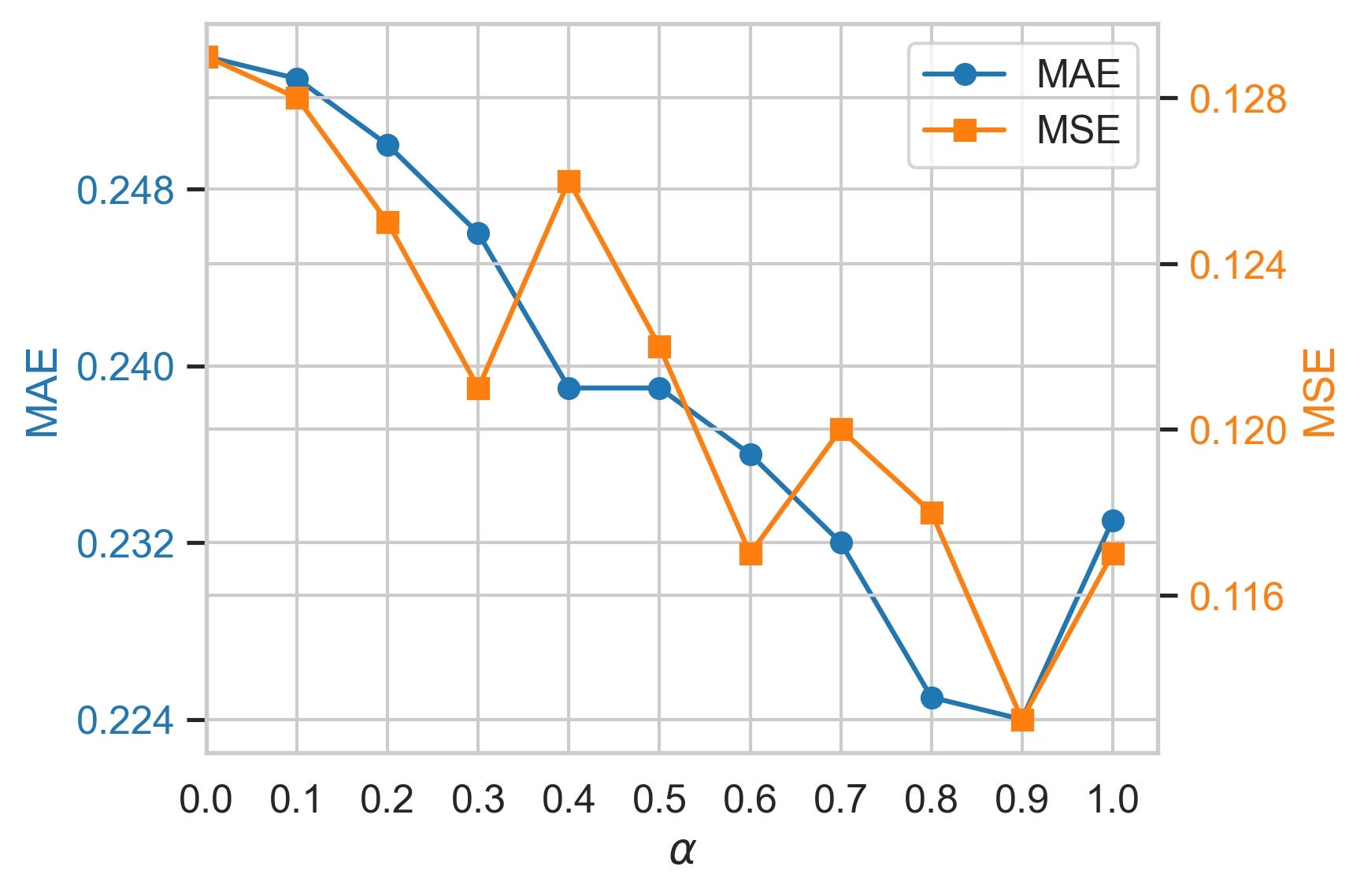}
        
        \label{curve_rmse_train}
    \end{subfigure}
    \hfill 
    \begin{subfigure}[b]{0.23\textwidth}
    \caption{StemGNN for AIR-BJ}
        \centering
        \includegraphics[width=\textwidth]{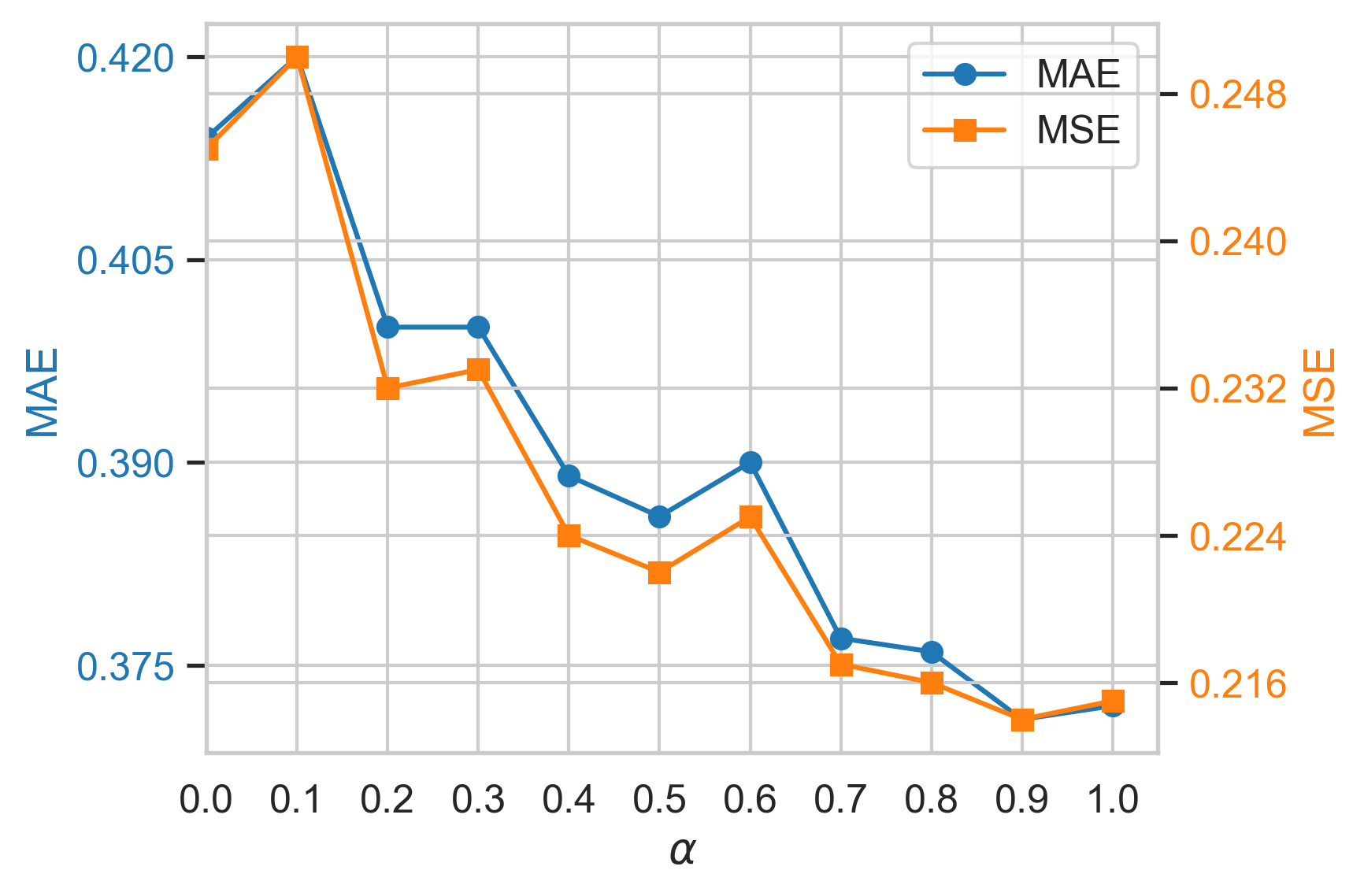}
        
        \label{Alpha}
    \end{subfigure}

    
    \label{alpha}
\end{figure}

In this section, we explore the role of the hyperparameter $\alpha$ in our framework. As illustrated in Figure~\ref{alpha}, the model performance generally improves as $\alpha$ increases, although a slight fluctuation in error metrics may occur toward the upper end of the range. We observed that the optimal value of $\alpha$ varies across different datasets; for instance, the optimal $\alpha$ for the AIR-BJ dataset is approximately $0.9$. These findings indicate that unifying supervision signals from both the time and frequency domains effectively enhances predictive accuracy, further validating the necessity of integrating frequency-domain constraints.

\subsubsection{Graph Construction}

Table~\ref{tab:adj_ablation} presents our ablation study on three distinct graph construction strategies, revealing that the structural quality of the adjacency matrix significantly impacts forecasting accuracy. We find that graph construction based on simple physical connectivity fails to capture complex latent topologies, thereby hindering effective spatial decorrelation during training. Crucially, the ability to accurately capture intricate spatial interdependencies is not merely a model configuration but the very foundation upon which effective spatio-temporal modeling is built; without a robust topological representation, the model remains blind to the essential relational patterns in the data. We also observe that the performance variations in our main results tend to coincide with specific structural properties, further highlighting the complex interplay between graph structure and spectral supervision. Consequently, investigating optimal, potentially dynamic graph construction methods represents a critical avenue for future research to further unlock the potential of spatio-temporal models.

\begin{table}[t]
\centering
\caption{Ablation study on adjacency matrix construction for SH-METRO dataset. Best results are in \textcolor{red}{red}.}
\label{tab:adj_ablation}
\footnotesize 
\setlength{\tabcolsep}{5pt}
\renewcommand{\arraystretch}{1.1}
\begin{tabular}{l rr rr rr}
\toprule
\multirow{2}{*}{Adjacency Matrix} & \multicolumn{2}{c}{DLinear} & \multicolumn{2}{c}{iTransformer} & \multicolumn{2}{c}{STID} \\
\cmidrule(lr){2-3} \cmidrule(lr){4-5} \cmidrule(lr){6-7}
& MAE & RMSE & MAE & RMSE & MAE & RMSE \\
\midrule
Similarity  & \textcolor{red}{106.233} & \textcolor{red}{236.295} & 66.596 & \textcolor{red}{160.290} & \textcolor{red}{81.525} & \textcolor{red}{187.214} \\
Correlation & 115.952 & 262.763 & \textcolor{red}{64.459} & 169.892 & 86.055 & 195.356 \\
Connection  & 115.717 & 263.021 & 74.178 & 181.131 & 84.543 & 190.916 \\
\bottomrule
\end{tabular}

\vspace{1mm}
{\scriptsize\raggedright The Physical Connection Matrix ($P$) is constructed based on the realistic topology of the metro system to reflect direct station connectivity; the Similarity Score Matrix ($S$) quantifies the resemblance of evolutionary flow patterns between station pairs using Dynamic Time Warping (DTW); and the Correlation Ratio Matrix ($C$) models the directional flow interactions derived from historical Origin-Destination (OD) distributions. Please refer to the original paper for details \citep{liu2020physical}.\par}
\vspace{-5mm}  
\end{table}


\section{Conclusion and Future Work}
In this paper, we propose FreST Loss (Frequency-enhanced spatio-temporal Loss) to address the fundamental challenge of autocorrelation bias in spatio-temporal forecasting. While standard point-wise objectives like MSE implicitly assume conditional independence among future observations, FreST Loss transforms the optimization landscape into a joint frequency domain by composing \textit{FFT} and \textit{GFT}. This approach leverages the asymptotic independence of spectral components to effectively decorrelate complex spatio-temporal dependencies. Extensive experiments across six benchmarks demonstrate that FreST Loss is model-agnostic and consistently improves the performance of various state-of-the-art architectures, including STGNNs, Transformers, and linear/MLP models, confirming its robust applicability across diverse modeling paradigms.

Future research will focus on two critical dimensions to further unlock the potential of spectral supervision. First, we will investigate the impact of graph construction strategies, exploring how dynamic and data-driven latent topologies interact with spatial frequency alignment compared to static physical connections. Second, we aim to refine the integration of \textit{FFT}, \textit{GFT}, and \textit{JFT} components. This includes developing advanced adaptive fusion mechanisms and training-phase-dependent weighting schemes to better disentangle and prioritize coupled spectral signals, while also extending the framework to handle irregularly sampled and non-stationary spatio-temporal data, advancing the frontier of predictive modeling.

\section{Ethical Use of Data and Informed Consent}

This work exclusively uses publicly available benchmark datasets (METR-LA, PEMS, NYC-BIKE, AIR-BJ, AIR-GZ) that contain aggregated sensor readings without personally identifiable information. No human subjects experiments were conducted, and IRB approval was not required. 

\bibliographystyle{ACM-Reference-Format}
\bibliography{refs}










\end{document}